\documentclass[10pt,journal,compsoc]{IEEEtran}
\ifCLASSOPTIONcompsoc
 \usepackage{cite}
\else
 \usepackage{cite}
\fi
\ifCLASSINFOpdf
\else
\fi
\usepackage{amsmath}
\usepackage{algorithmic}
\usepackage{graphicx}
\usepackage{subfigure}
\usepackage{booktabs} 
\usepackage{amsthm}
\usepackage{amssymb}
\usepackage{bm}
\usepackage{newtxmath}
\usepackage{xcolor}
\usepackage{xspace}
\usepackage{bbm}
\usepackage{algorithm}
\usepackage{algorithmic}
\usepackage[nolist]{acronym}
\usepackage{hyperref}
\usepackage[shortlabels]{enumitem}

\usepackage{mathtools}
\usepackage{amsmath}
\usepackage{setspace}
\hypersetup{
 colorlinks=false,
 linkcolor=black,
 filecolor=black, 
 urlcolor=blue,
 }

\usepackage{amsmath,amsfonts,bm}

\def\eqref#1{equation~\ref{#1}}
\def\Eqref#1{Equation~\ref{#1}}

\def\1{\bm{1}}

\def\vmu{{\bm{\mu}}}
\def\vtheta{{\bm{\theta}}}
\def\va{{\bm{a}}}

\def\ve{{\bm{e}}}
\def\vf{{\bm{f}}}
\def\vg{{\bm{g}}}

\def\vl{{\bm{l}}}
\def\vm{{\bm{m}}}

\def\vp{{\bm{p}}}
\def\vq{{\bm{q}}}

\def\vs{{\bm{s}}}

\def\vu{{\bm{u}}}
\def\vv{{\bm{v}}}

\def\vx{{\bm{x}}}

\def\mI{{\bm{I}}}
\def\mJ{{\bm{J}}}
\def\mK{{\bm{K}}}

\def\mM{{\bm{M}}}

\def\mR{{\bm{R}}}

\def\mT{{\bm{T}}}

\DeclareMathAlphabet{\mathsfit}{\encodingdefault}{\sfdefault}{m}{sl}
\SetMathAlphabet{\mathsfit}{bold}{\encodingdefault}{\sfdefault}{bx}{n}

\DeclareMathOperator*{\argmin}{arg\,min}

\theoremstyle{plain}

\newtheorem*{theorem*}{Theorem}

\def\vmu{{\bm{\mu}}}

\newcommand{\tran}{^\intercal}
\newcommand{\inv}{^{\text{-}1}}

\def\vtau{{\bm{\tau}}}
\def\vmu{{\bm{\mu}}}

\begin{document}
%
\title{A Differentiable Newton-Euler Algorithm \\ for Real-World Robotics}
%
%
%
%

\author{Michael~Lutter, Johannes Silberbauer, Joe Watson and~Jan~Peters
\IEEEcompsocitemizethanks{
\IEEEcompsocthanksitem M. Lutter, J. Silberbauer, J. Watson and J. Peters are with the Intelligent Autonomous Systems group with the Computer Science Department, Technical University of Darmstadt, Darmstadt, Germany. \protect\\
E-mail: michael@robot-learning.de
}
\thanks{
}}

%
%

\markboth{}%
{Lutter \MakeLowercase{\textit{et al.}}: [...]}
%



\IEEEtitleabstractindextext{%
\begin{abstract}
Obtaining dynamics models is essential for robotics to achieve accurate model-based controllers and simulators for planning. The dynamics models are typically obtained using model specification of the manufacturer or simple numerical methods such as linear regression. However, this approach does not guarantee physically plausible parameters and can only be applied to kinematic chains consisting of rigid bodies.
%
In this article, we describe a differentiable simulator that can be used to identify the system parameters of real-world mechanical systems with complex friction models, holonomic as well as non-holonomic constraints. To guarantee physically consistent parameters, we utilize virtual parameters and gradient-based optimization. The described Differentiable Newton-Euler Algorithm (DiffNEA) can be applied to a class of dynamical systems and guarantees physically plausible predictions.
%
The extensive experimental evaluation shows, that the proposed model learning approach learns accurate dynamics models of systems with complex friction and non-holonomic constraints. Especially in the offline reinforcement learning experiments, the identified DiffNEA models excel. For the challenging ball in a cup task, these models solve the task using model-based offline reinforcement learning on the physical system. The black-box baselines fail on this task in simulation and on the physical system despite using more data for learning the model. 
\end{abstract}

}

\maketitle

\IEEEdisplaynontitleabstractindextext

%
\IEEEpeerreviewmaketitle

\begin{acronym}
\acro{hjb}[HJB]{Hamilton-Jacobi-Bellman}
\acro{hji}[HJI]{Hamilton-Jacobi-Isaacs}
\acro{sim2real}[Sim2Real]{simulation to real}
\acro{rl}[RL]{reinforcement learning}
\acro{cfvi}[cFVI]{Continuous Fitted Value Iteration}
\acro{rfvi}[rFVI]{Robust Fitted Value Iteration}
\acro{mlp}[MLP]{Multi-Layer Perceptron}
\acro{promp}[ProMP]{Probabilistic Movement Primitive}
\end{acronym}

\IEEEraisesectionheading{\section{Introduction}\label{sec:introduction}}

%
%
%
%
\IEEEPARstart{T}{he} identification of dynamical systems from data is a powerful tool in robotics \cite{aastrom1971system}.
Learned analytic models may be used for control synthesis and can be utilized for gravitational and inertial compensation \cite{spong2020robot}. 
Moreover, when used as simulators, they can be used to reduce the sample complexity of data-driven control methods such as \ac{rl}~\cite{deisenroth2011pilco, chua2018deep, hafner2019dream, lutter2021learning}.
For these control applications, where out-of-distribution prediction is typically required, the ability to generalize beyond the acquired data is critical.
Any modeling error may be exploited by a controller, and such exploitation may result in catastrophic system failure.
To ensure sufficient out-of-sample generalization, the model's hypothesis space is an important consideration.
Ideally, this space should be defined such that only plausible trajectories, that are physically consistent and have bounded energy, are generated. 

\medskip
\noindent Standard black-box models such as deep networks or Gaussian processes have a broad hypothesis space and can model arbitrary dynamical systems with high fidelity. Therefore, these black-box models have been frequently used for model learning for control~\cite{nguyen2011survey, nguyen2008computed, nguyen2009model} as well as model-based reinforcement learning~\cite{deisenroth2011pilco, chua2018deep, hafner2019dream, lutter2021learning}. However, one disadvantage of these black-box models is that the learned approximation is valid locally and not physically consistent. Therefore, these models do not generalize out-of-distribution and can generate trajectories with unbounded energy. 
Only white-box models \cite{khosla1985parameter, mukerjee1985self, atkeson1986estimation, gautier1986identification, ting2006bayesian, traversaro2016identification, wensing2017linear, sutanto2020encoding, geist2021structured}, which are derived from first principles and infer the physical parameters of the analytic equations of motion, can guarantee out-of-sample generalization as these models are valid globally. 
While these combinations of physics with data-driven learning can obtain more robust representations, the usage of white-box models commonly reduces model accuracy compared to black-box methods and has been mainly applied to rigid body systems with holonomic constraints~\cite{khosla1985parameter, mukerjee1985self, atkeson1986estimation, gautier1986identification, ting2006bayesian, traversaro2016identification, wensing2017linear, sutanto2020encoding, geist2021structured}. The white-box models obtain a lower accuracy as the Newtonian-, Lagrangian- and Hamiltonian mechanics used to derive the equations of motion typically cannot describe the complex nonlinear phenomena of friction with high fidelity. Therefore, these models commonly underfit for physical systems.

\medskip 
\noindent In this article, we show that learning the system parameters by utilizing differentiable simulators can alleviate many shortcomings of classical white-box models. When the equations of motion are differentiable, we show that one can include complex friction models and apply this approach to systems with non-holonomic constraints. We present the Differentiable Newton-Euler Algorithm~(DiffNEA)~\cite{lutter2020differentiable, lutter2021differentiable} and combine this approach with various actuator models ranging from white-box friction models to deep network friction models. In the experiments, we apply this differentiable simulator to the identification of multi-body robot dynamics. We benchmark these models on the simulated and physical Furuta Pendulum and Cartpole. The experiments show that DiffNEA models with energy-bounded black-box actuator models yield non-divergent trajectories and improve the predicted rollouts. Other black-box models that do not guarantee passive friction models are susceptible to learn dynamics that generate energy and lead to divergence. 

\medskip \noindent
For systems with non-holonomic constraints, we show that these constraints can be added to the optimization as additional penalties. Solving the resulting optimization problem with gradient-based optimization enables the identification of the physical parameters of the non-holonomic constraints. In the experiments, we apply this technique to solve ball in a cup with offline model-based reinforcement learning on the physical system. This task is especially challenging as one needs to learn the string dynamics and obtain a model that cannot be exploited by the reinforcement learning agent. The DiffNEA model can learn an accurate and robust model that cannot be exploited. The optimal solution obtained using the DiffNEA model can be transferred to the physical system and successfully achieves the task for multiple string lengths. In contrast, black-box models can be easily exploited by the RL agent and fail when transferred to the physical system. 

\subsection{Contribution}
The main contribution of this article is to show that differentiable simulators can alleviate many problems of classical white-box models. Using the proposed approach, white-box models are not limited to rigid-body systems but can include separate friction models and non-holonomic constraints. To demonstrate that differentiable simulators can learn the system parameters of complex physical systems and highlight the advantages of this approach:
\medskip
\begin{enumerate} [wide=0pt]
 \item We describe the DiffNEA that utilizes differentiable simulation, gradient-based optimization, and virtual parameters to infer physically plausible system parameters of the rigid bodies, the constraints, and the friction models. \vspace{0.25em}
 \item We perform an extensive experimental evaluation to measure the performance of the DiffNEA models on the physical system where the assumptions of rigid-body systems do not hold. The performed evaluation shows that the differentiable simulator can be applied to physical systems and outperform the black-box models when out-of-distribution generalization is essential. 
\end{enumerate} 

\subsection{Outline}
The article is structured as follows. First, we summarize the different approaches to learning dynamics models for robotics (Section~\ref{sec:dynamics_models}). Section~\ref{sec:diffnea} describes DiffNEA. The subsequent experimental section (Section~\ref{sec:experiments}) summarizes the experimental setup and presents the results. Finally, Section~\ref{sec:conclusion} discusses the obtained results and summarizes the results of the article.

\section{Dynamics Model Representations} \label{sec:dynamics_models}
\noindent Model learning, or system identification \cite{aastrom1971system}, aims to infer the parameters $\vtheta$ of the system dynamics from data containing the system state $\vx$ and the control signal $\vtau$.
In the continuous time case the dynamics are described by
\begin{gather}
 \ddot{\vx} = \vf(\vx, \dot{\vx}, \vtau; \vtheta).
\end{gather}
The optimal parameters $\vtheta^{*}$ are commonly obtained by minimizing the error of the forward or inverse dynamics model,
\begin{align} \label{eq:loss}
 \vtheta^{*}_{\text{for}} &= \argmin_{\vtheta}
 \sum_{i=0}^{N} \| \ddot{\vx}_i - \hat{\vf} \left(\vx_i, \dot{\vx}_i, \vtau_i; \vtheta \right)\|^2,  \\
 \vtheta^{*}_{\text{inv}} &= \argmin_{\vtheta}
 \sum_{i=0}^{N} \| \vtau_i - \hat{\vf}^{\text{-}1}\left(\vx_i, \dot{\vx}_i, \ddot{\vx}_i; \vtheta \right)\|^2.
\end{align}
Depending on the chosen representation for $\vf$, the model hypotheses spaces and the optimization method changes. Generally one can differentiate between black-box models and white-box models. Recently there are also have been various gray-box models~\cite{nguyen2010using, lutter2020differentiable, hwangbo2019learning, allevato2020tunenet} to bridge the gap and combine parts of both categories. 

\subsection{Black-box Models} 
\noindent These models use generic function approximators $f$ and the corresponding abstract parameters $\vtheta$ to represent the dynamics. Within the model learning literature many function approximation technique has been applied, including locally linear models \cite{Atkeson_AIR_1998, schaal2002scalable}, Gaussian processes \cite{kocijan2004gaussian, nguyen2009model, nguyen2010using}, deep- \cite{hafner2019dream, lutter2021learning, chua2018deep} and graph networks \cite{sanchez2020learning, sanchez2018graph}. These approximators can fit arbitrary and complex dynamics with high fidelity but have an undefined behavior outside the training distribution and might be physically unplausible even in the training domain. Due to the local nature of the representation, the behavior is only well defined on the training domain and hence the learned models do not extrapolate well. Furthermore, these models can learn implausible systems violating fundamental physics laws such as energy conservation. Furthermore, these models can only learn either the forward or inverse models and are not interpretable. Therefore, only the input-output relation can be computed but no additional information such as the system energies or momentum. Only recently deep networks have been augmented with knowledge from physics to constrain network representations to be physically plausible on the training domain and interpretable \cite{lutter2018deep, Lutter2019Energy, greydanus2019hamiltonian, gupta2019general, cranmer2020lagrangian, saemundsson20variational, zhong2019symplectic}. However, the behavior outside the training domains remains unknown. 

\subsection{White-box Models}
\noindent These models use the analytical equations of motions to formalize the hypotheses space of $\vf$ and the interpretable physical parameters such as mass, inertia, or length as parameters $\vtheta$. Commonly the equations of motion are derived using either Newtonian, Lagrangian, or Hamiltonian mechanics. Therefore, white-box models are limited to describe the phenomena incorporated within the equations of motions but generalize to unseen state regions as the parameters are global. The classical approach to obtain the system parameters is by measuring these properties of the disassembled system or estimating them using the CAD software~\cite{albu2002regelung}. Instead of measuring the parameters, four different groups concurrently proposed to estimate these parameters from data~\cite{khosla1985parameter, mukerjee1985self, atkeson1986estimation, gautier1986identification}. These papers showed that the recursive Newton-Euler algorithm (RNEA) \cite{Featherstone2007rigid} for rigid-body chain manipulators simplifies to a linear model. Therefore, the parameters can be inferred using linear least squares. Since then, this approach of data-driven system identification has been widely used and improved~\cite{yoshida2000verification, mata2005dynamic, gautier2013identification, gautier2013identification_2, sousa2014physical, traversaro2016identification, wensing2017linear, sutanto2020encoding, ayusawa2011real}. 

\medskip\noindent
The resulting parameters must not necessarily be physically plausible as constraints between the parameters exist. For example, the inertia matrix contained in $\vtheta^{*}$ must be a positive definite matrix and fulfill the triangle inequality~\cite{ting2006bayesian}. Since then, various parameterizations for the physical parameters have been proposed to enforce these constraints through the virtual parameters.
Various reparameterizations \cite{traversaro2016identification, wensing2017linear, ayusawa2011real, sutanto2020encoding} were proposed to guarantee physically plausible inertia matrices. Using these virtual parameters, the optimization does not simplify to linear regression but can be solved by unconstrained gradient-based optimization and is guaranteed to preserve physical plausibility. Commonly this non-linear optimization technique was solved using sequential quadratic programming where the derivatives were approximated using finite differences~\cite{traversaro2016identification, wensing2017linear, ayusawa2011real}. 

\medskip\noindent
The previous approaches assumed that the equations of motion are differentiable and the optimization objective smooth. Otherwise, the system parameters cannot be inferred using gradient-based optimization. To avoid these assumptions, multiple authors~\cite{ramos2019bayessim, barcelos2020disco, muratore2021neural} have proposed likelihood-free inference methods to identify the posterior distribution over the parameters. In this case, the equations of motions can be a black box that only needs to be evaluated. For standard physics engines, e.g., PyBullet~\cite{coumans2018} and MuJoCo~\cite{6386109}, this assumption is favorable as these simulators are not differentiable and a black-box.
Similarly to likelihood-free inference parameter identification for black-box simulators, Jiang et. al. \cite{jiang2021simgan} proposed to obtain the distribution of the physical parameters using reinforcement learning. These approaches are commonly applied to domain randomization as these techniques automatically tune the randomization for the specific physical system. Therefore, the resulting policies are not too conservative but achieve the simulation to reality transfer. 

\medskip\noindent
In this article, we assume that the equations of motion are known and differentiable. Therefore, we use a gradient-based approach to infer the optimal parameters. We utilize the recent availability of automatic differentiation (AD)~\cite{Rall81} frameworks to compute the gradient w.r.t. to the parameters analytically using backpropagation. Furthermore, we use the virtual parametrization proposed by \cite{traversaro2016identification, wensing2017linear} to guarantee that the physical parameters are plausible without enforcing additional constraints within the optimization.

\begin{algorithm}[t]
\caption{The Articulated Body Algorithm computing the joint accelerations for a kinematic tree in terms of Lie algebra \cite{kim2012lie}.}
\label{alg:articulated-rigid-body}
\setstretch{1.4}
\begin{algorithmic}
\STATE {\bfseries Input:} Joint Position $\vq$, Joint Velocity $\dot{\vq}$, Torque $\vtau$ 
\STATE {\bfseries Output:} Joint Acceleration $\ddot{\vq}_{i}$
\FOR{$i = 1$ \TO $n$}
\STATE Compute forward kinematics along the chain\;
\STATE $\bar{\vv}_i = Ad_{T^{-1}_{\lambda, i}}\bar{\vv}_{\lambda} {+} \vs_i \dot{\vq}_i$
\STATE $\bar{\bm{\eta}}_i = ad_{\bar{\vv}_i} \vs_i \: \dot{\vq}_i$
\ENDFOR
\FOR{$i = n$ \TO $1$}
\STATE Compute lumped inertia for each body
\STATE $\hat{\bar{\mM}}_{i:n} {=} \bar{\mM}_i {+} \sum_{k \in \mu} Ad_{T_{i,k}^{-1}}^* \bar{\bm{\Pi}}_k Ad_{T_{i,k}^{-1}}$
\STATE Compute bias forces for each link
\STATE $\bar{\vf}_{b, i} = - ad_{\bar{\vv}_i}^* \bar{\mM}_{i:n} \bar{\vv}_i {+} \sum_{k \in \mu} Ad_{T_{i,k}^{-1}}^* \Big( \bar{\vf}_{b, k} {+} \bar{\bm{\beta}}_k \Big)$
\STATE $\bar{\Psi}_i = \big(\vs_i^{\top} \: \hat{\bar{\mM}}_{i:n} \: \vs_i\big)^{-1}$
\STATE  $\bar{\bm{\Pi}}_i = \hat{\bar{\mM}}_{i:n} {-} \hat{\bar{\mM}}_{i:n} \vs_i \bar{\Psi}_i \vs_i^{\top} \hat{\bar{\mM}}_{i:n}$
\STATE $\bar{\bm{\beta}}_i = \hat{\bar{\mM}}_{i:n}
\Big(\bar{\bm{\eta}}_i{+}\vs_i \bar{\Psi}_i \Big( u_i {-} \vs_i^{\top} \Big( \hat{\bar{\mM}}_{i:n} \bar{\bm{\eta}}_i {+} \bar{\vf}_{b, i} \Big) \Big) \Big)$
\ENDFOR
\FOR{$i = 1$ \TO $n$}
\STATE Compute acceleration using Newton-Euler \;
\STATE $\ddot{\vq}_i = \bar{\Psi}_i \Big( \vtau_i {-} \vs_i^{\top} \Big(\hat{\bar{\mM}}_{i:n} \Big( Ad_{T_{\lambda, i}^{-1}} \bar{\va}_{\lambda} {+} \bar{\eta}_i \Big) {-} \bar{\vf}_{b, i} \Big)\Big)$\;
\STATE $\bar{\va}_i = Ad_{T_{\lambda, i}^{-1}} \bar{\va}_{\lambda} {+} \vs_i \ddot{\vq}_i {+} \bar{\bm{\eta}}_i$\;
\STATE $\bar{\vf}_i = \hat{\bar{\mM}}_{i:n} \bar{\va}_i {+} \bar{\vf}_{b, i}$\;
\ENDFOR
\end{algorithmic}
\end{algorithm}

\subsection{Differentiable Simulators}
As we assume that the equations of motion are differentiable, the proposed system identification approach using gradient-based optimization is closely related to differentiable simulators. Recently various approaches to differentiable simulation~\cite{de2018end, hu2019difftaichi, geilinger2020add, degrave2019differentiable, heiden2019interactive, heiden2021disect, werling2021fast, heiden2020augmenting} has been proposed to enable system identification via gradients and policy optimization using backprop through time~\cite{miller1995neural}.
%
The main problem of differentiable simulators is differentiating through the contact and the constraint force solver computing $\vf_{c}$. The equations of motion of articulated rigid bodies without contacts only require linear algebra operations and hence, are differentiable by default. To differentiate through the contacts, various approaches have been proposed. For example, Belbute-Peres et. al.~\cite{de2018end} and Heiden et. al.~\cite{heiden2020augmenting} describe a method to differentiate through the common LCP solver of simulators, Degrave et. al.~\cite{degrave2019differentiable} utilize impulse-based contacts based on constraint violation to enable gradient computation using automatic differentiation and avoid solving the LCP, Geilinger et. al. \cite{geilinger2020add} describe a smoothed contact model to enable the differentiation and Hu et al. \cite{hu2019difftaichi} describe a continuous collision resolution approach to improve the gradient computation. A great in-depth description of the different approaches to differentiable simulation and their advantages is provided in~\cite{werling2021fast}.

\medskip\noindent
In this article, we mainly only simulate articulated rigid bodies that are differentiable by default. For ball in a cup, the constraint forces originating from the non-holonomic constraints can be computed in closed form. Therefore, we can easily differentiate through these analytic expressions of the constraint forces and do not need to rely on more complex approaches presented in the literature.

\section{Differentiable  Newton-Euler Algorithm}
\label{sec:diffnea} 
\noindent In this section, we describe the used differentiable system identification technique that is based on the Newton-Euler algorithm in terms of the elegant Lie algebra formulation \cite{kim2012lie}. First, we describe the identification approach for systems with holonomic constraints, i.e., kinematic trees (Section~\ref{sec:holonomic}). Afterwards, we describe the virtual parameters that enable physically plausible system parameters (Section~\ref{sec:virtual}) and extend them to systems with non-holonomic constraints (Section~\ref{sec:non_holonomic}). Finally, the different actuators are introduced in Section~\ref{sec:actuators}.

\subsection{Rigid-Body Physics \& Holonomic Constraints} \label{sec:holonomic}
\noindent
For rigid-body systems with holonomic constraints, the system dynamics can be expressed analytically in maximal coordinates $\vx$, i.e., task space, and reduced coordinates $\vq$, i.e., joint space.
If expressed using maximal coordinates, the dynamics is a constrained problem with the holonomic constraints $g(\cdot)$.
For the reduced coordinates, the dynamics are reparametrized such that the constraints are always fulfilled and the dynamics are unconstrained.
Mathematically these dynamics are described by
\begin{gather}
 \ddot{\vx} = f(\vx, \dot{\vx}, \vtau; \vtheta) \hspace{15pt}
 \text{s.t.} \hspace{15pt} 
 g(\vx; \vtheta) = 0, \\
 \Rightarrow
 \ddot{\vq} = f(\vq, \dot{\vq}, \vtau; \vtheta).
\end{gather}
For model learning of such systems one commonly exploits the reduced coordinate formulation and minimizes the squared loss of the forward or inverse model.
For kinematic trees the forward dynamics $\vf(\cdot)$ can be easily computed using the articulated body algorithm (ABA, Algorithm~\ref{alg:articulated-rigid-body}) and the inverse dynamics $\vf\inv(\cdot)$ via the recursive Newton-Euler algorithm (RNEA, Algorithm~\ref{alg:rnea})~\cite{Featherstone2007rigid}. Both algorithms are inherently differentiable and one can solve the optimization problem of \Eqref{eq:loss} using backpropagation.

\medskip
\noindent In this implementation, we use the Lie formulations of ABA and RNEA \cite{kim2012lie} for compact and intuitive compared to the initial derivations by \cite{atkeson1986estimation, Featherstone2007rigid}.
ABA and RNEA propagate velocities and accelerations from the kinematic root to the leaves and the forces and impulses from the leaves to the root. This propagation along the chain can be easily expressed in Lie algebra by
\begin{align}
 \bar{\vv}_{j} &= \text{Ad}_{\mT_{j,i}}\bar{\vv}_i, &
 \bar{\va}_{j} &= \text{Ad}_{\mT_{j,i}}\bar{\va}_i, \\
 \bar{\vl}_{j} &= \text{Ad}^{\top}_{\mT_{j,i}}\bar{\vl}_i, &
 \bar{\vf}_{j} &= \text{Ad}^{\top}_{\mT_{j,i}}\bar{\vf}_i.
\end{align}
with the generalized velocities $\bar{\vv}$, accelerations $\bar{\va}$, forces $\bar{\vf}$, momentum $\bar{\vl}$ and the adjoint transform $\text{Ad}_{\mT_{j,i}}$ from the $i$th to the $j$th link.
The generalized entities noted by $\bar{.}$ combine the linear and rotational components, e.g., $\bar{\vv} = \left[\vv, \bm{\omega} \right]$ with the linear velocity $\vv$ and the rotational velocity $\bm{\omega}$.
The Newton-Euler equation is described by 
\begin{gather*}
 \bar{\vf}_{\text{net}} = 
 \bar{\mM}\bar{\va} - \text{ad}^*_{\bar{\vv}}\bar{\mM}\bar{\vv},
 \\
 \text{ad}^*_{\bar{\vv}} = 
 \begin{bmatrix}
 [\bm{\omega}] & \mathbf{0} \\
 [\vv] & [\bm{\omega}]
 \end{bmatrix}, 
 \hspace{10pt}
 \bar{\mM} = 
 \begin{bmatrix}
 \mJ & m[\vp_m] \\
 m[\vp_m]^{\top} & m\mI
 \end{bmatrix},
\end{gather*}
with the inertia matrix $\mJ$, the link mass $m$, the center of mass offset $\vp_m$. Combining this message passing with the  Newton-Euler equation enables a compact formulation of RNEA and ABA.

\begin{algorithm}[t]
\caption{The Recursive Newton-Euler Algorithm computing the input torque for a kinematic tree in terms of Lie algebra \cite{kim2012lie}.}
\label{alg:rnea}
\setstretch{1.4}
\begin{algorithmic}
\STATE {\bfseries Input:} Position $\vq$, Velocity $\dot{\vq}$, Acceleration $\ddot{\vq}$
\STATE {\bfseries Output:} Torque $\vtau$ 
\FOR{$i = 1$ \TO $n$}
\STATE // Forward Kinematics\;
\STATE $\bar{\vv}_i = Ad_{T^{-1}_{\lambda, i}}\bar{\vv}_{\lambda} {+} \vs_i \dot{\vq}_i$ \;
\STATE $\bar{\va}_i = Ad_{T^{-1}_{\lambda, i}}\bar{\va}_{\lambda} + ad_{\bar{\vv}_i} \vs_i \: \dot{\vq}_i + \vs_i \: \ddot{\vq}_i$
\ENDFOR
\FOR{$i = n$ \TO $1$}
\STATE // Newton-Euler Equation \;
\STATE $\bar{\vf}_i = \bar{\mM}_i\bar{\va}_i - ad_{\bar{\vv}_i}^* \bar{\mM}_i \bar{\vv}_i + \sum_{k \in \mu} Ad_{T_{i,k}^{-1}}^* \bar{\vf}_k$ \;
\STATE $\vtau_i = \vs_i\tran \bar{\vf}_i$
\ENDFOR
\end{algorithmic}
\end{algorithm}

\subsection{Virtual Physical Parameters} \label{sec:virtual}
To obtain the optimal parameters from data, one cannot simply minimize the mean squared error of the forward or inverse model as these parameters have additional constraints. For example, the transformation matrix $\mT$ must be a homogeneous transformation (i.e., $\mT \in \text{SE(3)}$), the inertias $\mJ$ must comply with the parallel axis theorem, and the masses $m$ must be positive. 
To obtain physically plausible parameters with gradient-based optimization without additional constraints, we reparametrize these physical parameters with \emph{virtual} parameters. These virtual parameters are \emph{unrestricted} and yield a physically plausible for all values \cite{traversaro2016identification, wensing2017linear, sutanto2020encoding}. For example, we optimize the square root of the mass to always obtain a positive mass. 

\subsubsection{Kinematics}\label{sec:kin} \noindent
The transformation $\mT(q)$ between two links depends on the link length, the joint position and the joint constraint connecting the two links. We decompose this transformation $\mT(q){=}\mT_{O} \mT_q(q)$ into a fixed transform $\mT_O$ and variable transform $\mT_q(q_i)$. The fixed transform describes the the distance and rotation between two joints and is parametrized by the translation vector $\vp_k$ and the RPY Euler angles $\vtheta_R{=}[\phi_x, \phi_y, \phi_z]^{\top}$. The transformation is then described by 
\begin{align}
 \mT_O &= 
 \begin{bmatrix}
 \mR_z(\phi_z) \mR_y(\phi_y) \mR_x(\phi_x) & \vp_k \\
 0 & 1
 \end{bmatrix},
\end{align}
where $\mR_a(\phi)$ denotes the rotation matrix corresponding to the rotation by $\phi$ about axis $a$ using the right-hand rule. Note that the rotation matrices about the elementary axis only depend on $\vtheta_R$ through arguments to trigonometric functions. Due to the periodic nature of those functions we obtain a desired unrestricted parameterization. The variable transform $\mT_q(q_i)$ describes the joint constraint and joint configuration. For all joints we assume that the variable link axis is aligned with the z-axis. Hence, the transformation matrix and joint motion vector for revolute joints ($\mT_{q_{r}},\vs_r$) and prismatic joints ($\mT_{q_{p}},\vs_p$) is described by
\begin{alignat*}{2}
 &\mT_{q_r} = 
 \begin{bmatrix}
 \mR_{z}(q) & \mathbf{0} \\
 \mathbf{0} & 1
 \end{bmatrix}, \hspace{25pt}
 \mT_{q_p} &&= 
 \begin{bmatrix}
 \mathbf{0} & q \: \ve_z\\
 \mathbf{0} & 1
 \end{bmatrix}, \\
 &\:\:\vs_r = \: \begin{bmatrix}
 \mathbf{0} & \ve_z
 \end{bmatrix}, \hspace{48pt}
 \vs_p &&= \begin{bmatrix}
 \ve_z & \mathbf{0}
 \end{bmatrix}, 
\end{alignat*}
with the $z$ axis unit vector $\ve_z$.
Technically, one can also use fixed joints with $\mT_q{=}\mathbf{I}$ but this simply yields an overparameterized model and is not necessary for describing the system dynamics. The complete kinematics parameters of a link are summarized as $\vtheta_K{=}\{\vtheta_R, \vp_k \}$.

\subsubsection{Inertias} \noindent
For physical correctness, the diagonal rotational inertia $\mJ_p{=}\text{diag}([J_x, J_y, J_z])$ at the body's CoM and around principal axes must be positive definite and need to conform with the triangle inequalities~\cite{traversaro2016identification}, i.e.,
\begin{equation*}
J_x \leq J_y + J_z , \quad 
J_y \leq J_x + J_z , \quad
J_z \leq J_x + J_y \;.
\end{equation*}
To allow an unbounded parameterization of the inertia matrix, we introduce the parameter vector
$\vtheta_L{=}[\theta_{\sqrt{L_1}},\theta_{\sqrt{L_2}},\theta_{\sqrt{L_3}}]\tran$, 
where $L_i$ represents the central second moments of mass of the density describing the mass distribution of the rigid body with respect to a principal axis frame. Then rotational inertia is described by
\begin{align*}
 \mJ_p = \text{diag}(
\theta_{\sqrt{L_2}}^2{+}\theta_{\sqrt{L_3}}^2, \:
\theta_{\sqrt{L_1}}^2{+}\theta_{\sqrt{L_3}}^2, \:
\theta_{\sqrt{L_1}}^2{+}\theta_{\sqrt{L_2}}^2
).
\end{align*}
The rotational inertia is then mapped to the link coordinate frame using the parallel axis theorem described by 
\begin{equation}
 \mJ = \mR_J \mJ_p \mR_J\tran + m[\vp_m][\vp_m],
\end{equation}
with the link mass $m$ and the translation $\vp_m$ from the coordinate from to the CoM. 
The fixed affine transformation uses the same parameterization as described in \ref{sec:kin}. 
The mass of the rigid body is parameterized by $\theta_{\sqrt{m}}$ where $m{=}\theta_{\sqrt{m}}^2$.
Given the dynamics parameters $\vtheta_{I}{=}\{\vtheta_L, \theta_{\sqrt{m}}, \vtheta_J, \vp_m\}$ for each link, the inertia in the desired frame using as well as generalized inertia $\bar{\mM}$ can be computed.

\subsection{Rigid-Body Physics \& non-holonomic Constraints} \label{sec:non_holonomic}
\noindent
For a mechanical system with non-holonomic constraints, the system dynamics cannot be expressed in terms of unconstrained equations with reduced coordinates. For the system
\begin{align*} \label{eq:non-holonomic}
 \ddot{\vx} = f(\vx, \dot{\vx}, \vu; \vtheta) \hspace{10pt}
 \text{s.t.}\hspace{10pt}
 h(\vx; \vtheta) \leq 0,\hspace{10pt}
 g(\vx, \dot{\vx}; \vtheta) = 0,
\end{align*}
the constraints are non-holonomic as $h(\cdot)$ is an inequality constraint and $g(\cdot)$ depends on the velocity.
Inextensible strings are an example for systems with inequality constraint, while the bicycle is a system with velocity dependent constraints.
For such systems, one cannot optimize the unconstrained problem directly but must identify parameters that explain the data and adhere to the constraints. 

\medskip
\noindent The dynamics of the constrained rigid body system can be described by the Newton-Euler equation,
\begin{align}
\bar{\vf}_{\text{net}} &= \bar{\vf}_{g} + \bar{\vf}_{c} + \bar{\vf}_{\vu} = \bar{\mM}\bar{\va} - \text{ad}^*_{\bar{\vv}}\bar{\mM}\bar{\vv}, \\
\Rightarrow \bar{\va} &= \bar{\mM}\inv \left(\bar{\vf}_{g} + \bar{\vf}_{c} + \bar{\vf}_{\vu} + \text{ad}^*_{\bar{\vv}}\bar{\mM}\bar{\vv} \right),
\end{align}
where the net force $\bar{\vf}_{\text{net}}$ contains the gravitational force $\bar{\vf}_{g}$, the constraint force $\bar{\vf}_{c}$ and the control force $\bar{\vf}_{\vu}$.
If one can differentiate the constraint solver computing the constraint force w.r.t. to the parameters,
one can identify the parameters $\vtheta$ via gradient descent. This optimization problem can be described by
\begin{align*}
\min_{\vtheta}
\sum_{i=0}^{N} \Big| \: \bar{\va}_i {\,-\,} \bar{\mM}^{\text{-}1}_{\vtheta} \Big(\bar{\vf}_{g}(\vtheta)
{\,+\,} \bar{\vf}_{c}(\bar{\vx}_i, \bar{\vv}_i; \vtheta) {\,+\,} \bar{\vf}_{\vu} {\,+\,} \text{ad}^*_{\bar{\vv}_i}\bar{\mM}(\vtheta)\bar{\vv}_i \Big)\: \Big|^2.
\end{align*}
For the inequality constraint, one can reframe it as an easier equality constraint, by passing the function through a ReLU nonlinearity $\sigma$, so $g(\vx; \vtheta){\,=\,}\sigma(h(\vx; \vtheta)){\,=\,}0$.
From a practical perspective, the softplus nonlinearity provides a soft relaxation of the nonlinearity for smoother optimization.
Since this equality constraint should always be enforced, we can utilize our dynamics to ensure this on the derivative level, so
$g(\cdot) {\,=\,} \dot{g}(\cdot) {\,=\,} \ddot{g}(\cdot) {\,=\,} 0$ for the whole trajectory.
With this augmentation, the constraint may now be expressed as $\vg(\vx,\dot{\vx};\vtheta){\,=\,}\mathbf{0}$.
The complete loss is described by 
\begin{align*}
\min_{\vtheta} \sum_{i=0}^{N} \| \bar{\va}_i - \vf(\bar{\vx}_i, \bar{\vv}_i, \bar{\vu}_i; \vtheta)\|^2  + \vg(\vx_i,\dot{\vx}_i;\vtheta)^{\top} \bm{\Lambda}\: \vg(\vx_i,\dot{\vx}_i;\vtheta),
\end{align*}
with the penalty weighting $\bm{\Lambda} = \text{diag}\left( \lambda_{g}, \lambda_{\dot{g}}, \lambda_{\ddot{g}} \right)$.

\begin{figure*}[t]
\centering
\includegraphics[width=\linewidth]{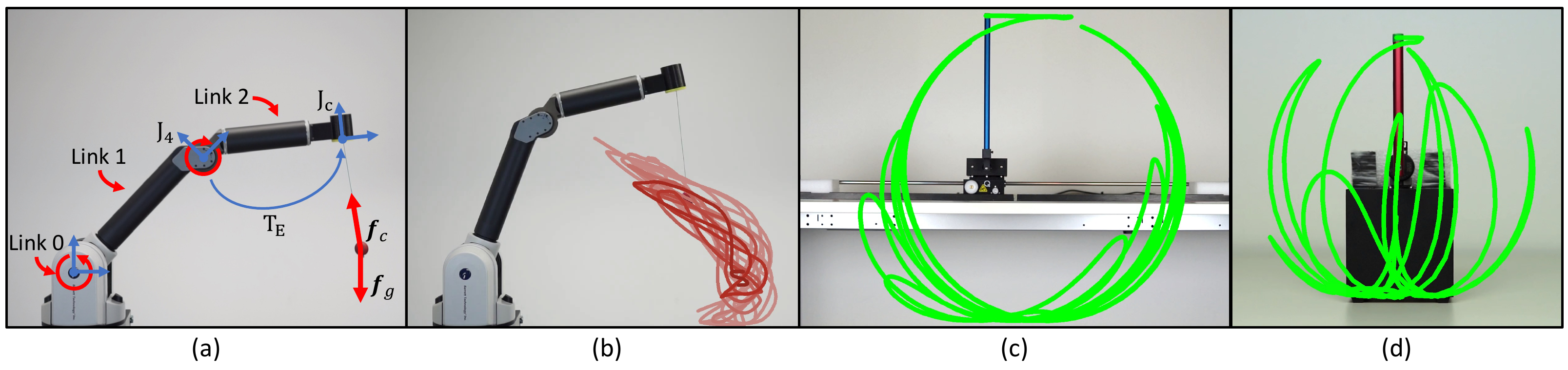} \\ 
\caption{(a) The identified dynamics (red) and kinematic (blue) parameter of the Barrett WAM for the ball in a cup task. (b) Exploration data for the DiffNEA model to infer $T_{E}$ and the string length. (c) The Quanser cartpole perfoming an swing-up movement. (d) The Quanser Furuta pendulum performing the swing-up movement. The green trajectory highlights the performed movements of the under-actuated systems.}
 \label{fig:title}
\end{figure*}

\subsection{Actuator Models} \label{sec:actuators} 
All described models simulate rigid body systems that conserve the energy and assume that the actuators apply the desired torque $\vtau_d$, i.e., $\vtau = \vtau_d$. For physical systems, this representation does not capture reality with sufficient fidelity as these systems dissipate energy via friction and the actuators are not ideal, i.e., $\vtau \approx \vtau_d$. For most articulated kinematic trees, the friction of the joints and actuators dominates compared to the air resistance. To incorporate these phenomena into white-box models, we augment the rigid-body body simulator with an actuator model that incorporates joint friction and transforms the desired torque into the applied torque, i.e., $\vtau = f(\vtau_d, \vq, \dot{\vq})$.

\medskip\noindent
This actuation model can either be white-box model relying on existing friction models or black-box models. We define five different joint independent actuator models that cover the complete spectrum from white-box to black-box model, 
\begin{align*}
 \text{Viscous:}& &\vtau &= \vtau_d {-} \vmu_v {\odot} \dot{\vq}, \\[0.1em]
 \text{Stribeck:}& &\vtau &= \vtau_d {-} \sigma(\dot{\vq}) {\odot} \big(\vf_s {+} \vf_d {\odot} \exp\big({-}\nu_s \dot{\vq}^2 \big)\big) {-} \vmu_v {\odot} \dot{\vq},\\[0.1em]
 \text{NN-Friction:}& &\vtau &= \vtau_d {-} \sigma(\dot{\vq}) {\odot} \| f_{\text{NN}}(\vq,\dot{\vq}; \psi) \|_1 ,\\[0.1em]
 \text{NN-Residual:}& &\vtau &= \vtau_d {+} f_{\text{NN}}(\vq,\dot{\vq}; \psi), \\[0.1em]
 \text{FF-NN:}& &\vtau &= f_{\text{NN}}(\vtau_d, \vq,\dot{\vq}; \psi),
\end{align*}
with the deep networks weights $\psi$ and the elementwise multiplication $\odot$. The Viscous and Stribeck actuator models are white-box models that have been proposed within the literature~\cite{olsson1998friction, albu2002regelung, bona2005friction, wahrburg2018motor}. These models assume that the the motor is ideal as well as that the friction is additive, independent for each joint and does not depend on the robot state. The resulting system dynamics is guaranteed to yield a stable uncontrolled system as these models dissipate energy, i.e., $\dot{E} = \bm{\tau}^{\top} \dot{\vq} \leq 0$. The black-box alternative to these friction model is the NN friction model. This friction model also assumes that applied torque is ideal and guarantees that the system is passive. However, in this case the frictional torque depends on the robot configuration and can represent more complex shapes. To not only model the actuator friction that dissipates energy, the NN residual and FF-NN actuator model can model arbitrary actuator dynamics.

\medskip \noindent
To learn the parameters of the actuator models, we add the parameters of the models to the gradient-based optimization. We regularize the training of the actuator model by adding penalties that prevent the actuator model dynamics to dominate the system dynamics. Similar optimization procedures have been used in \cite{lutter2020differentiable, hwangbo2019learning, allevato2020tunenet} to train existing grey-box models that combine deep networks and analytic models. 

\section{Experiments} \label{sec:experiments}
In the experiments, we apply DiffNEA to three physical systems. We evaluate the long-term predictions of the learned forward models and test whether the models can be used for reinforcement learning. Using this evaluation, we want to answer the following questions:

\medskip\noindent
\textbf{Q1:} Can DiffNEA models learn accurate dynamics models on the physical system even when the systems include stiction or non-holonomic constraints? 

\medskip\noindent
\textbf{Q2:} When do DiffNEA models significantly outperform black-box models on the physical system?

\subsection{Experimental Setup} 
To answer these questions we perform two separate experiments. In the first, we identify the parameters of two under-actuated systems on different datasets and compare their simulated trajectories. The main objective of this experiment is to compare a large number of different system identification approaches and characterize their similarities and differences. In the second, we learn the ball in a cup dynamics and use the model for offline reinforcement learning. In this experiment, we want to test whether the learned models can be exploited by the reinforcement learning agent. In the following, we describe the used systems, model variations, and tasks in detail.

\subsubsection{Physical Dynamical Systems}
We apply the DiffNEA to identify the parameters of the cartpole, the Furuta pendulum, and the Barrett WAM. For all physical systems, only the joint position is directly observed. The velocities and accelerations are computed using finite differences and low-pass filters. These filters are applied offline to use zero-phase shift filters that do not cause a phase shift within the observations. 

\medskip \noindent
\textbf{Barrett WAM.} The Barrett WAM (Figure~\ref{fig:title}a \& \ref{fig:title}b) consists of four fully actuated degrees of freedom controlled via torque control with $500$Hz. The actuators are back-driveable and consist of cable drives with low gear ratios enabling fast accelerations. The joint angles sensing is on the motor-side. Therefore, any deviation between the motor position and joint position due to the slack of the cables cannot be observed.

\medskip \noindent
\textbf{Cartpole.} The physical cartpole (Figure~\ref{fig:title}c) is an under-actuated system manufactured by Quanser~\cite{quanser}. The pendulum is passive and the cart is voltage controlled with up to $500$Hz. The linear actuator consists of a plastic cogwheel drive with high stiction. 

\medskip \noindent
\textbf{Furuta Pendulum.} 
The physical Furuta pendulum (Figure~\ref{fig:title}d) is an under-actuated system manufactured by Quanser~\cite{quanser}. Instead of the linear actuator of the cartpole, the Furuta pendulum has an actuated revolute joint and a passive pendulum. The revolute joint is voltage controlled with up to $500$Hz. The main challenge of this system is the small masses and length scales that requires a sensitive controller.

\subsubsection{Parametric Dynamics Models}
For the evaluation, we compare black-box models to three different instantiations of the DiffNEA model family. Each of these models is combined with the applicable actuation models and different model initialization. We differentiate between two initialization strategies, without prior and with prior. Without prior means that the link parameters are initialized randomly. With prior means that the parameters are initialized with the known values given by the manufacturer. This differentiation enables us to evaluate the impact of good initialization for white-box models. All models are continuous-time models and we use the Runge-Kutta 4 method~(RK4) for integration. All non-linear optimization problems are solved using gradient descent and ADAM~\cite{kingma2014adam}. 

\medskip\noindent
\textbf{Newton-Euler Algorithm.}
The Newton-Euler Algorithm (NEA) model assumes knowledge of the kinematic chain and the kinematics parameters $\vtheta_K$ and only learns the inertial parameter $\vtheta_I$. These parameters are learned by linear regression. This model learning approach was concurrently introduced by \cite{khosla1985parameter, mukerjee1985self, atkeson1986estimation, gautier1986identification}. Due to the linear regression, this model cannot be augmented with the different actuation models.

\medskip\noindent
\textbf{No-Kin Differential  Newton-Euler Algorithm.}
The no-Kin DiffNEA model assumes knowledge of the kinematic tree but no knowledge of the kinematics $\vtheta_K$ or inertial parameters $\vtheta_I$. The link parameters are learned by minimizing the squared loss of the forward dynamics. 

\medskip\noindent
\textbf{Differential  Newton-Euler Algorithm.}
The DiffNEA model assumes knowledge of the kinematic chain and the kinematics parameters $\vtheta_K$ and only learns the inertial parameters $\vtheta_I$. These parameters are learned by minimizing the squared loss of the forward dynamics.

\medskip\noindent
\textbf{Feed-Forward Neural Network.}
The black-box model learning baseline is a feed-forward neural network (FF-NN). This network is a continuous-time model and predicts the joint acceleration.

\subsubsection{Evaluation Tasks}
The different models are applied to two different tasks, trajectory prediction, and offline reinforcement learning.

\medskip \noindent
\textbf{Trajectory Prediction.} To evaluate the learned forward models, we use these models to predict the trajectory from an initial state $\vx_0$ and action sequence $\vu_{0:T}$. The predicted trajectory is compared to the trajectory of the true model. To evaluate the impact of the dataset, we evaluate the performance on three different datasets with different levels of complexity. The uniform dataset is obtained by sampling joint positions, velocities, and torques from a uniform distribution and computing the acceleration with the true analytic forward dynamics. The simulated trajectory dataset is generated by simulating the ideal system with viscous friction and small state and action noise. The real system dataset is generated on the physical system by applying an energy controller that repeatedly swings up the pendulum and lets the pendulum fall without actuation. 

\begin{figure*}[t]
 \centering
 \includegraphics[width=\textwidth]{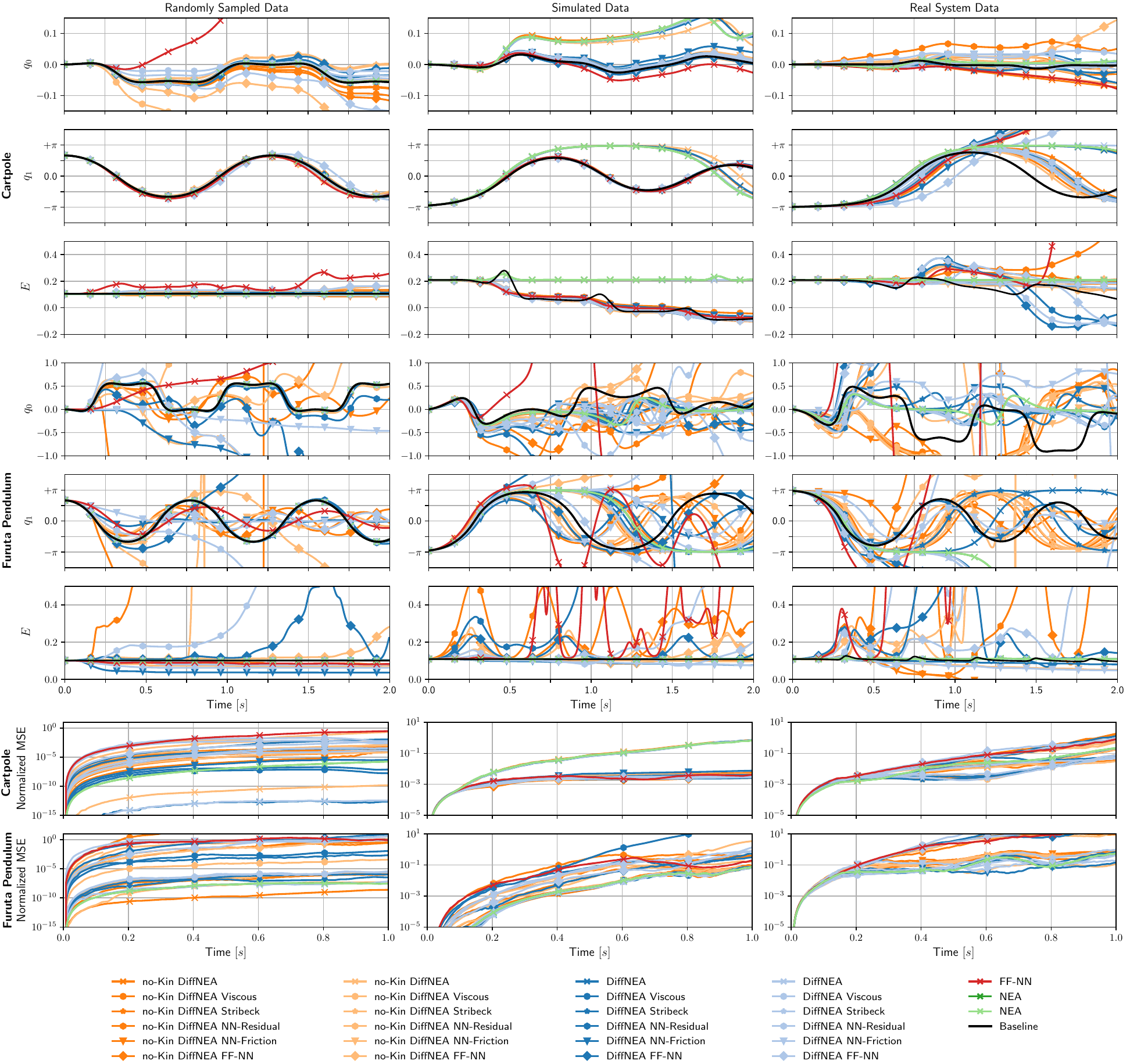}
 \caption{Qualitative model comparison of the 26 different models performing forward roll-outs on the Cartpole and Furuta Pendulum. The roll-outs start from the starting state and are computed with 250Hz sampling frequency and integrated with RK4. The models are trained on three different datasets ranging from uniformly sampled and ideal observations (i.e., simulated data from uniform sampling) to trajectory data of noisy observation from the physical system.}
 \label{fig:rollout}
\end{figure*}

\medskip \noindent
\textbf{Offline Reinforcement Learning.} To solve the ball in cup task, we use model-based offline reinforcement learning. In this setting, one learns a model from a fixed dataset of arbitrary experience and uses this model to learn the optimal policy \cite{levine2020offline, Lange2012}. Hence, the agent is bound to a dataset and cannot explore the environment. More specifically, we use episodic relative entropy policy search~(eREPS)~\cite{peters2010relative} with an additional KL-divergence constraint on the maximum likelihood policy update \cite{ploeger2020high} and parameter exploration \cite{deisenroth2013survey} to optimize a distribution over trajectories described using a \ac{promp}~\cite{paraschos2013probabilistic, paraschos2018using}.

\medskip \noindent
For the manipulator identification the robot executes a $40$s high-acceleration sinusoidal joint trajectory (Figure \ref{fig:title} b).
For the string model identification, the robot executes a $40$s slow cosine joint trajectories to induce ball oscillation without contact with the manipulator (Figure \ref{fig:title} c). The ball trajectories are averaged over five trajectories to reduce the variance of the measurement noise.
The training data does not contain swing-up motions and, hence the model must extrapolate to achieve the accurate simulation of the swing-up.
The total dataset used for offline RL contains only $4$ minutes of data. To simplify the task for the deep networks, the training data consists of the original training data plus all data generated by the DiffNEA model. Therefore, the network training data contains successful BiC tasks. 

\medskip \noindent
The dense episodic reward is inspired by the potential of an electric dipole and augmented with regularizing penalties for joint positions and velocities.
The complete reward is defined as
\begin{align*}
 R {\,=\,} \exp \left( \frac{1}{2}\max_t \psi_t {\,+\,} \frac{1}{2}\psi_N \right)
 {\,-\,} \frac{1}{N} \sum_{i=0}^{N} \lambda_{\vq} \| \vq_i {-} \vq_0 \|^2_2 {\,+\,} \lambda_{\dot{\vq}} \| \dot{\vq}_i \|^2_2,
\end{align*}
with $\psi_t = \Delta_t \tran \hat{\vm}(\vq_t) /\left( \Delta_t \tran \Delta_t + \epsilon \right)$ and the normal vector of the end-effector frame
$\hat{\vm}$ which depends on joint configuration $\vq_t$. 
For the DiffNEA model, the \emph{predicted} end-effector frame is used during policy optimization.
Therefore, the policy is optimized using the reward computed in the approximated model. The black-box models uses the true reward, rather than the reward bootstrapped from the learned model.

\medskip \noindent
For the DiffNEA models, the robot manipulator is modeled as a rigid-body chain using reduced coordinates. The ball is modeled via a constrained particle simulation with an inequality constraint.
Both models are interfaced via the task space movement of the robot after the last joint.
The manipulator model predicts the task-space movement after the last joint. The string model transforms this movement to the end-effector frame via $\mT_{E}$ (Figure \ref{fig:title} a), computes the constraint force $\vf_c$ and the ball acceleration $\ddot{\vx}_B$.
Mathematically this model is described by
\begin{gather}
 \ddot{\vx}_B = \frac{1}{m_B} \left( \vf_g + \vf_c \right), \\ 
 g(\vx; \vtheta_S) = \sigma(\| \vx_B - \mT_{E} \: \vx_{J_4} \|^2_2 - r^2) = 0,
\end{gather}
where $\vx_B$ is the ball position, $\vx_{J_4}$ the position of the last joint and $r$ the string length. In the following we will abbreviate $\vx_B -\mT_{E} \: \vx_{J_4} = \Delta$ and the cup position by $\mT_{E} \: \vx_{J_4} = \vx_C$. The constraint force can be computed analytically with the principle of virtual work and is described by
\begin{equation}
\begin{gathered}
 \vf_c(\vtheta_S) = - m_B \: \sigma'(z) \: \frac{\Delta\tran \vg - \: \Delta\tran \ddot{\mathbf{x}}_C + \: \dot{\Delta}\tran \dot{\Delta}}{\Delta\tran \Delta + \delta}, 
\end{gathered}
\end{equation}
with $ z = \| \Delta \|_2 - r,$ and the gravitational vector $\vg$. When simulating the system, we set $\ddot{g} = -\mK_p g - \mK_d \dot{g} \leq 0$ to avoid constraint violations and add friction to the ball for numerical stability. This closed form constraint force is differentiable and hence one does not need to incorporate any special differentiable simulation variants.

\medskip \noindent For the black-box models, a feedforward network (FF-NN) and a long short-term memory network (LSTM) \cite{hochreiter1997long} is used. The networks model only the string dynamics and receive the task space movement of the last joint and the ball movement as input and predict the ball acceleration, i.e.,
$\ddot{\vx}_B = f(\vx_{J_4}, \dot{\vx}_{J_4}, \ddot{\vx}_{J_4}, \vx_{B}, \dot{\vx}_{B})$. 

\subsection{Experimental Results}
This section presents the experimental observations of the trajectory prediction and offline \ac{rl} experiment. After each experiment, the conclusion from each experiment is stated. 

\begin{figure*}[t]
 \centering
 \includegraphics[width=\textwidth]{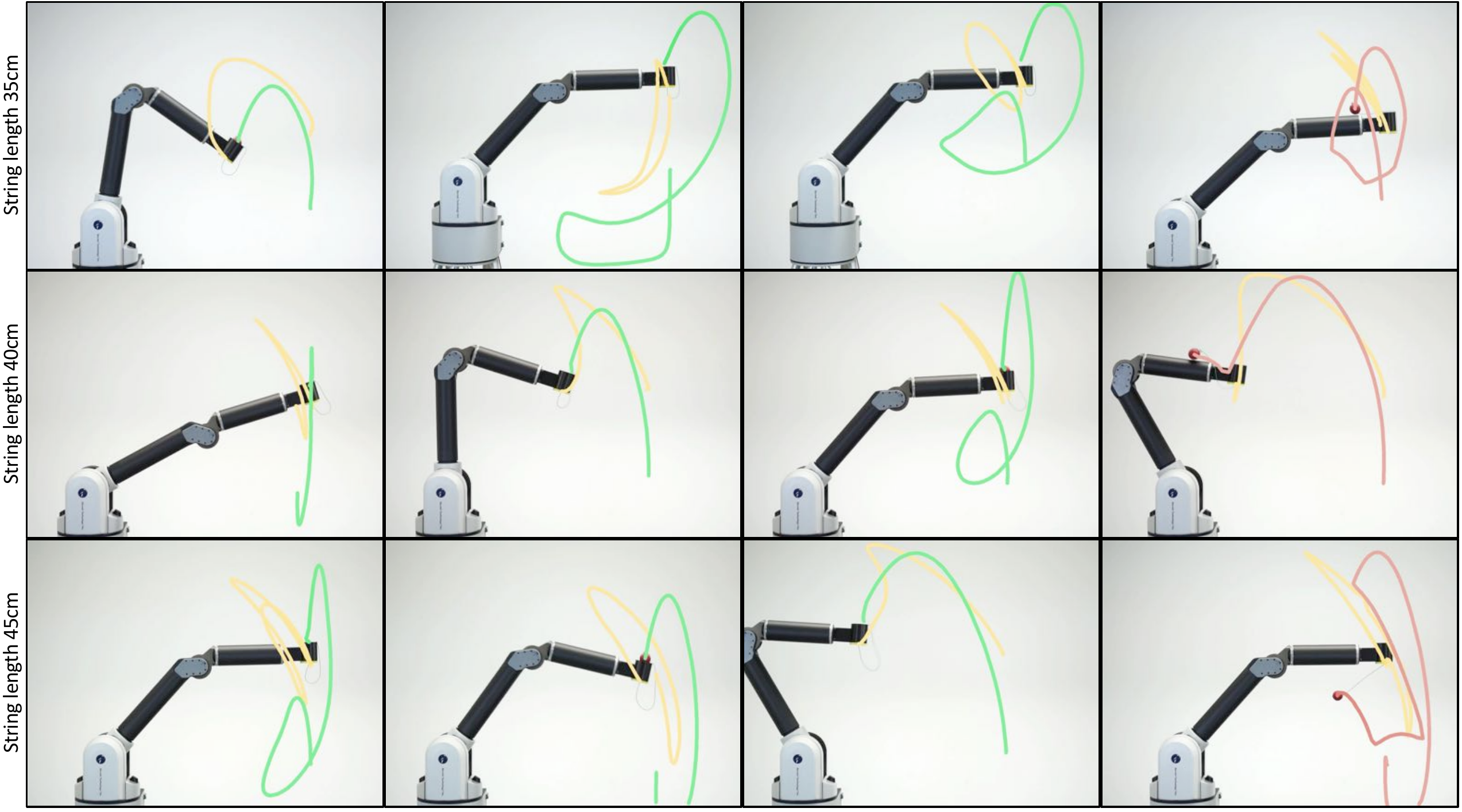}
 \caption{Three different successful swing-ups for the three different string lengths using the DiffNEA model with eREPS for offline model-based reinforcement learning. This approach can learn different swing-ups from just 4 minutes of data, while all tested black-box models fail at the task. The different solutions are learned using different seeds. The unsuccessful trials of the DiffNEA model nearly solve the BiC tasks but the ball bounces off the cup or arm. 
 }
 \label{fig:movements}
\end{figure*} 

\subsubsection{Trajectory Prediction}
The predicted trajectories and the normalized mean squared error are shown in Figure~\ref{fig:rollout}. The overall performance depends on the system as well as the dataset. The numerically sensitive conditioning of the Furuta Pendulum causes all models to be worse on all datasets and model classes. Conversely, the magnitude of the physical parameters of the cartpole makes the identification and long-term prediction simpler. Regarding the datasets, the overall forward prediction performance decreases with dataset complexity. For example, the uniform dataset yields perfect DiffNEA models with very small errors. For the simulated and real trajectory datasets, the prediction error deteriorates. Regarding the different models, no clear best system identification approach is apparent. One can only observe significant differences between the different model classes. 

\medskip \noindent
\textbf{Energy Bounded vs. Energy Unbounded Models.}
The energy-conserving models (i.e., no-actuator model) perform well when the observed system conserves energy or nearly conserves energy (e.g., Furuta Pendulum). If the systems dissipate a lot of energy, the model prediction degrades. For example, the energy-conserving models do not obtain an accurate prediction for the \emph{simulated} cartpole that contains high viscous friction. For the Furuta pendulum, the friction is negligible, hence the energy-conserving models perform well. 
%
The energy-bounded models (i.e., Viscous, Stribeck, and NN-Friction actuator model) are the best performing models of this benchmark. The learned models yield non-diverging trajectories and can model systems with and without friction. The NN-Friction actuator achieves good performance by exploiting its black-box flexibility. Despite its expressiveness, the predicted trajectories do not diverge. 

\medskip \noindent
The energy-unbounded models (i.e., FF-NN and NN-Residual actuator model and the black-box model), frequently learn to inject energy into the system even for perfect sensor measurements. For example, the black-box deep network increases the system energy of the cartpole even for the simulated uniformly sampled data. For the more challenging trajectory datasets, all energy-unbounded models predict trajectories that increase in energy during simulation without actuation. Many of these models also generate divergent trajectories during simulation, which leave the training domain. 

\medskip \noindent
\textbf{Learned Kinematics vs. Known Kinematics.}
One interesting observation is that the DiffNEA model with \emph{unknown} kinematics (no-Kin DiffNEA) performs comparable to the DiffNEA model with \emph{known} kinematics (DiffNEA, NEA), demonstrating the kinematics and dynamics can be learned jointly. However, as the reinforcement learning experiment shows, this observation does not apply to more complex systems. For the ball in a cup experiment, the kinematic structure must be incorporated to learn a good model of the task. Otherwise one can only learn an accurate model of the WAM but not the cup and string dynamics. 

\medskip \noindent
\textbf{Model Initialization.}
In the simulation experiments, no clear difference between models with and without prior initialization is visible. Evaluating the identified physical parameters also yields no clear improvement of the initialization with prior, e.g., even for unreasonably large physical parameters, the identified parameter with a prior was not necessarily smaller. Therefore, we conclude that the initialization with the best-known parameters does not necessarily improve model performance. 

\medskip \noindent
\textbf{Conclusion.}
The experiments show that all models perform comparably the same in terms of the prediction horizon. This horizon mostly depends on the system characteristics and dataset. The horizon is shorter for the sensitive Furuta pendulum and longer for the cartpole. The horizon is longer for the simulated dataset with uniform state distribution compared to the trajectory datasets. The differences between the models are not significant. The main difference between these models is the behavior beyond this prediction horizon. The energy unbounded models frequently yield diverging trajectories while the energy bounded models do not diverge from the training domain. Therefore, we prefer the NN-friction model. This actuator model combines high model capacity with little assumptions on the friction model and yields non-diverging trajectories. 

\begin{table*}[t]
\caption{Offline reinforcement learning results for the ball in a cup task, across both simulation and the physical system. Length refers to the string length in centimeters. Repeatability is reported for the best performing reinforcement learning seed. The repeatability of the simulated system is not stated as the simulator is deterministic.}
\label{table:results}
  \begin{sc}
  \resizebox{\textwidth}{!}{%
  \begin{tabular}{r cccc cccc}
  	\toprule
  	& \multicolumn{4}{c}{simulation} & \multicolumn{4}{c}{physical system} \\
  	\cmidrule(lr){2-5} \cmidrule(lr){6-9}
	model & length & avg. reward & transferability & repeatability & length & avg. reward & transferability & repeatability \\
    \midrule
    lstm & 40cm & 0.92 $\pm$ 0.37 & 0\% & - & 40cm  & 0.91 $\pm$ 0.56 & 0\% & 0\% \\
    ff-nn & 40cm & 0.86 $\pm$ 0.35 & 0\%  & - & 40cm & 1.46 $\pm$ 0.78 & 0\% & 0\% \\
    Nominal & 40cm & 2.45 $\pm$ 1.15 & \textbf{64\%} & - & 40cm & 1.41 $\pm$ 0.45 & 0\% & 0\% \\
    diffnea & 40cm & \textbf{2.73 $\pm$ 1.64} & 52\%  & -
                             & 40cm & \textbf{1.77 $\pm$ 0.74} & \textbf{60\%} & 90\% \\
                         &&&&& 35cm & 1.58 $\pm$ 0.15 & 30\% & 70\%\\
                         &&&&& 45cm & 1.74 $\pm$ 0.71 & \textbf{60\%} & \textbf{100\%}\\
  \bottomrule
  \end{tabular}
 }
 \end{sc}
\end{table*}

\begin{figure}[t]
 \centering
 \includegraphics[width=\columnwidth]{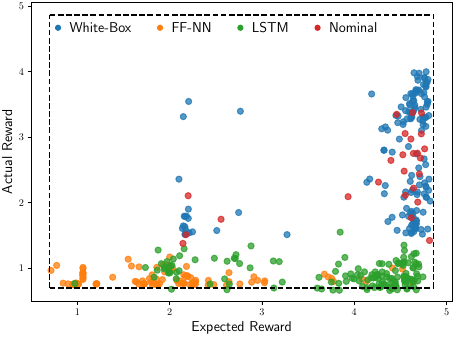}
 \caption{Comparison of the expected reward and the actual reward on the MuJoCo simulator for the LSTM, the feed-forward neural network (FF-NN) as well as the nominal and learned DiffNEA model. The learned and nominal model achieves a comparable performance and solves the BiC swing-up for multiple seeds. Neither the LSTM nor the FF-NN achieve a single successful swing-up despite being repeated with 50 different seeds and using all the data generated by the DiffNEA models.}
 \label{fig:results}
\end{figure}

\subsubsection{Offline Reinforcement Learning}
For this experiment, we only use the DiffNEA model with known kinematics of the Barrett WAM up to the last joint. The end-effector transformation from the last joint to the cup end-effector is learned. Learning the kinematics and the dynamics simultaneously did not yield accurate models as the joint optimization has too much ambiguity 
\footnote{Videos of the experiments are available at \hfill

\href{https://sites.google.com/view/ball-in-a-cup-in-4-minutes/}{https://sites.google.com/view/ball-in-a-cup-in-4-minutes/}}.

\medskip
\noindent \textbf{Simulation Results.}
The simulation experiments test the models with idealized observations from MuJoCo \cite{6386109} and enable a quantitative comparison across many seeds. For each model representation, 15 different learned models are evaluated with 150 seeds for the MFRL. The average statistics of the best ten reinforcement learning seeds are summarized in Table \ref{table:results} and the expected versus obtained reward is shown in Figure \ref{fig:results}.

\medskip \noindent
The DiffNEA model can learn the BiC swing-up for every tested model. The transferability to the MuJoCo simulator depends on the specific seed, as the problem contains many different local solutions and only some solutions are robust to slight model variations. The MuJoCo simulator is different from the DiffNEA model as MuJoCo simulates the string as a chain of multiple small rigid bodies. The performance of the learned DiffNEA is comparable to the performance of the DiffNEA model with the nominal values. 

\medskip \noindent
The FF-NN and LSTM black-box models do not learn a single successful movement that transfers to the physical system despite using ten different models, 150 different seeds, additional data that includes swing-ups, and observes the real instead of the imagined reward. These learned models cannot stabilize the ball beneath the cup. The ball immediately diverges to a physically unfeasible region. The attached videos compare the real (red) vs. imagined (yellow) ball trajectories. Within the impossible region, the policy optimizer exploits the random dynamics where the ball teleports into the cup. Therefore, the policy optimizers converge to random movements. 

\medskip \noindent \textbf{Real-Robot Results.}
On the physical Barrett WAM we evaluate 50 seeds per model. A selection of the trials using the learned DiffNEA model is shown in Figure~\ref{fig:movements}. The average statistics of the best ten seeds are summarized in Table \ref{table:results}. 

\medskip \noindent 
The DiffNEA model is capable of solving BiC using offline MBRL for all three string lengths. This approach obtains very different solutions that transfer to the physical system. Some solutions contain multiple pre-swings which show the quality of the model for long-planning horizons. The best movements achieve the task repeatedly. Solutions that do not transfer to the system, perform feasible movements where the ball bounces off the cup rim. The nominal DiffNEA model with the measured arm and string parameters does not achieve a successful swing-up. The ball always overshoots and bounces off the robot arm for this model. 

\medskip \noindent 
Similar to the simulation experiments, none of the tested black-box models achieve the BiC swing-up despite using more data and the true rewards during planning. Especially the FF-NN model converges to random policies, which result in ball movements that do not even closely resemble a potential swing-up. The convergence to very different movements shows that the models contain multiple shortcuts capable of teleporting the imagined ball into the cup. 

\medskip \noindent
\textbf{Conclusion.} The offline \ac{rl} experiment shows that the DiffNEA model excels when generalization is required and diverging from the training domain yields spurious solutions. The DiffNEA models achieve the task while black-box models fail. The learned DiffNEA models are not perfect. Hence, not all solutions obtained by the \ac{rl} agent achieve the task when transferred. The transfer rate could be improved by incorporating robustness within the \ac{rl} agent, e.g., domain randomization or optimizing the worst-case reward given an adversary controlling the dynamics.

\section{Conclusion} \label{sec:conclusion}
The Differentiable Newton-Euler Algorithm~(DiffNEA) learns physically consistent parameters for dissipative mechanical systems with holonomic and non-holonomic constraints. For this approach, we combine differentiable simulation, gradient-based optimization, and virtual parameters to infer physically plausible system parameters of the rigid bodies, the constraints, and the friction models. Therefore, this approach generalizes existing white-box models to a wider class of dynamical systems.

\medskip\noindent
The extensive experiments showed that this model learning technique can learn dynamics models of physical systems with friction and non-holonomic constraints. The learned models can be used for trajectory prediction and reinforcement learning. The obtained DiffNEA models learn more accurate dynamics models than standard black-box models such as deep networks. However, the accuracy of these DiffNEA models is not necessarily orders of magnitude better than the black-box models under realistic circumstances. Only when model assumptions are valid, the training data is uniformly sampled from the complete state domain and ideal joint observations are obtained, the DiffNEA model learns near-perfect models. Already when the training data is obtained using trajectories, the accuracy of DiffNEA models degrades significantly. 

\medskip\noindent
The main advantage of the DiffNEA models is the worst-case behavior and generalization. While black-box models commonly diverge from the training domain, DiffNEA models always yield physically consistent predictions. The long-term predictions of these models might not be highly accurate but cannot be exploited. Therefore, these models excel for \ac{rl}, where these characteristics are especially important. The offline \ac{rl} experiment shows that the DiffNEA model solves the ball in a cup task. The black-box models are not able to solve the task as the agent easily exploits the model and obtains a spurious solution. When transferred to the physical system, the robot performs a random trajectory.

\medskip\noindent
\textbf{Future Work.} To improve system identification techniques for white-box models, one needs to address the ambiguity of the physical parameters. The current approaches that minimize the 1-step error can yield very different parameter configurations as the physical parameters are over parametrized. For small parameters like friction, the 1-step loss is especially problematic as the impact of friction on the 1-step error is often negligible. However, small differences in the friction parameter lead to large errors for the long-term predictions. Therefore, an interesting future research direction is to optimize the system parameters on multiple timescales. Simply using the multi-step loss is not sufficient as backprop through time cannot be applied for very long horizons. Therefore, an interesting research direction for system identification is to leverage the advances of training generative models. For example, one could use a similar approach as generative adversarial networks~\cite{goodfellow2020generative} to identify parameters that yield accurate long-term predictions.

\ifCLASSOPTIONcompsoc
 \section*{Acknowledgments}
\else
 \section*{Acknowledgment}
\fi

\noindent
M. Lutter, J. Watson, and J. Peters received funding from the European Union’s Horizon 2020 research and innovation
program under grant agreement No \#640554 (SKILLS4ROBOTS). Furthermore, we want to thank the open-source projects NumPy~\cite{numpy} and PyTorch~\cite{pytorch}.

\ifCLASSOPTIONcaptionsoff
 \newpage
\fi



\bibliographystyle{IEEEtran}
\bibliography{IEEEabrv, refs}

\vfill



\end{document}